\documentclass[10pt,twocolumn,letterpaper]{article}

\usepackage{wacv}              %

\usepackage{graphicx}
\usepackage{amsmath}
\usepackage{amssymb}
\usepackage{booktabs}
\usepackage[symbol]{footmisc}

\usepackage[pagebackref,breaklinks,colorlinks]{hyperref}

\usepackage[capitalize]{cleveref}
\crefname{section}{Sec.}{Secs.}
\Crefname{section}{Section}{Sections}
\Crefname{table}{Table}{Tables}
\crefname{table}{Tab.}{Tabs.}

\def\wacvPaperID{1785} %
\def\confName{WACV}
\def\confYear{2025}

\usepackage{customization}
\usepackage{xcolor}
\usepackage{xspace}
\usepackage{smile_symbols}
\usepackage{graphicx}
\usepackage{enumitem}
\usepackage{bbding}
\usepackage{booktabs}
\usepackage{comment}
\usepackage{subcaption}
\usepackage{wrapfig}

\def\etal{\emph{et al}.}
\def\eg{\emph{e.g}.}
\def\ie{\emph{i.e}.}
\def\etc{\emph{etc}.}

\usepackage{amssymb}
\usepackage{pifont}
\definecolor{checkmark}{HTML}{40826D}
\definecolor{xmark}{HTML}{E62020}
\newcommand{\ua}[1]{{\textcolor{checkmark}{$\uparrow$ #1}}}
\newcommand{\da}[1]{{\textcolor{xmark}{$\downarrow$ #1}}}

\newcommand{\ours}{\texttt{3DzAL} \xspace}

\let\svthefootnote\thefootnote
\newcommand\freefootnote[1]{%
  \let\thefootnote\relax%
  \footnotetext{#1}%
  \let\thefootnote\svthefootnote%
}

\begin{document}

\title{Towards Zero-shot 3D Anomaly Localization}

\author{Yizhou Wang$^{1*}$ \quad \quad Kuan-Chuan Peng$^2$ \quad \quad Yun Fu$^1$\\
$^1$Northeastern University \quad \quad $^2$Mitsubishi Electric Research Laboratories \\
\small
\texttt{wyzjack990122@gmail.com \quad kpeng@merl.com \quad yunfu@ece.neu.edu}
\normalsize
}

\maketitle

\begin{abstract}
  3D anomaly detection and localization is of great significance for industrial inspection. Prior 3D anomaly detection and localization methods focus on the setting that the testing data share the same category as the training data which is normal. However, in real-world applications, the normal training data for the target 3D objects can be unavailable due to issues like data privacy or export control regulation. To tackle these challenges, we identify a new task -- zero-shot 3D anomaly detection and localization, where the training and testing classes do not overlap. To this end, we design \ours, a novel patch-level contrastive learning framework based on pseudo anomalies generated using the inductive bias from task-irrelevant 3D xyz data to learn more representative feature representations. Furthermore, we train a normalcy classifier network to classify the normal patches and pseudo anomalies and utilize the classification result jointly with feature distance to design anomaly scores. Instead of directly using the patch point clouds, we introduce adversarial perturbations to the input patch xyz data before feeding into the 3D normalcy classifier for the classification-based anomaly score. We show that \ours outperforms the state-of-the-art anomaly detection and localization performance. 
\end{abstract}

\freefootnote{*This work was done when Yizhou Wang was an intern at Mitsubishi Electric Research Laboratories.}

\section{Introduction}
\label{sec: intro}

\begin{figure}[tb]
  \centering
  \includegraphics[width=\linewidth]{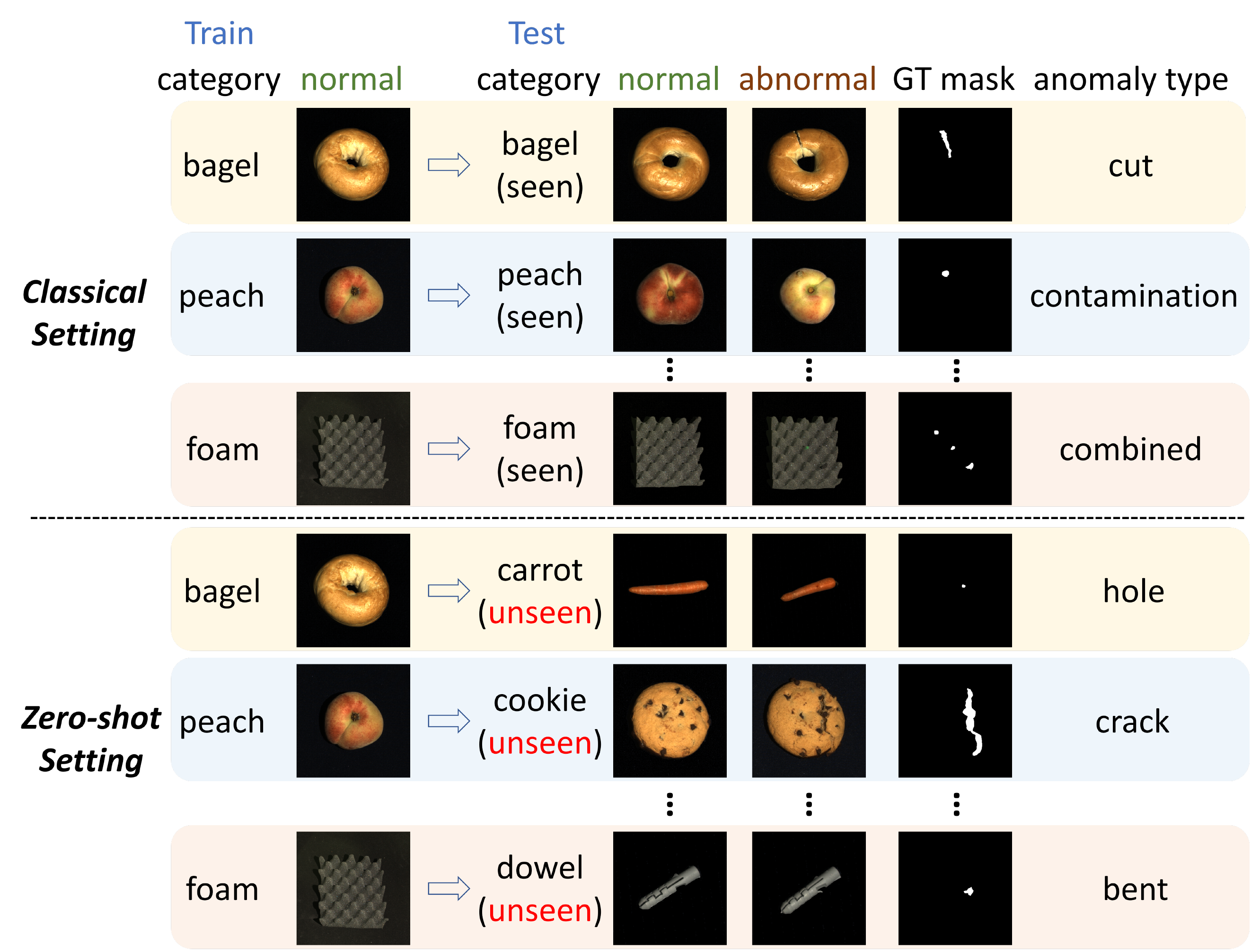}
  \caption{{\bf Problem overview.} Current 3D anomaly detection and localization works entail training on the normal data of one class and testing on the normal and abnormal data of the same class. We extend such setting by testing on other classes without the corresponding normal training data. This zero-shot setting is practical when such data are unavailable (\eg, due to data privacy, export control laws, \etc). GT denotes ground truth.}
  \label{fig: teaser}
\end{figure}
3D anomaly detection and localization methods have been highly demanded in real-world circumstances, including industrial inspection and autonomous driving~\cite{yu2020bdd100k,masuda2021toward,visapp22,xie2023iad,cao2023complementary,fan2021mpdnet,fan2020graph,wang2022making,wang2022self,wang2023towards,wang24tkde}. The main difference between 3D and 2D image anomaly detection and localization lies in that 3D data contain not only RGB information but also point location information~\cite{masuda2021toward,visapp22,cheraghian2022zero,cheraghian2020transductive}. Lots of shape anomalies of objects are readily identified as distinct sharp deformations from the point locations, in which cases, color information is less effective and the anomalies remain undetectable in top-down 2D views, as recognized in~\cite{Horwitz_2023_CVPR}. For instance, it is extraordinarily hard to identify a bent or cut location in a 2D image of a dowel, but such anomaly type can be very obvious in the 3D point cloud data. Recently, various 3D anomaly detection and localization methods have been introduced~\cite{Horwitz_2023_CVPR,zheng2022benchmarking,wan2022position,rudolph2023asymmetric,cao2023collaborative}. All these existing works concentrate on the setting that the testing data (including both normal and abnormal data) are from the same class as the training data. However, in real-world industrial 3D anomaly detection and localization applications, the normal training data of the target objects can be unavailable due to many possible reasons, \eg,  data privacy, export control regulations, \etc~Sometimes the normal data of the target objects on the client side are sensitive, and the client may not want to share the data but only want an anomaly detection and localization method that can perform well ``off-the-shelf." Therefore, a 3D anomaly detection and localization method able to generalize to unseen classes in the testing phase is needed.

\noindent \textbf{Problem Statement}. To address the aforesaid issues, we define a new problem in zero-shot 3D anomaly detection and localization, which involves identifying anomalies within a particular target class without any access to training data for that class or prior knowledge of its specific type of anomaly pattern. To be more specific, our goal is to localize abnormal locations in the 3D data in the target class's testing set, with no need of target class training data. Fig.~\ref{fig: teaser} illustrates this problem.

\noindent \textbf{Proposed framework}. To solve the aforesaid new problem, we propose a novel framework, namely ``3D zero-shot anomaly localization'' (\ours).
To achieve satisfactory zero-shot performance in 3D anomaly detection and localization, we add a learnable 3D feature extraction network on top of the 3D FPFH~\cite{rusu2009fast} features and encourage the learned features to be complementary to the features captured in FPFH.
To regularize the 3D feature extraction network, we use a patch-level pseudo anomaly-based contrastive learning scheme. We propose a pseudo anomaly generation module to synthesize anomalies since the training data only include the normal data without any abnormal data. When designing the pseudo anomaly generation module, we find that a randomly initialized and untrained CNN is able to locate the places of interest in three-dimensional point cloud data in its feature activation maps, \ie, if we feed the three-dimensional point cloud data as the input of a random CNN, the highly activated areas in this CNN's feature activation maps usually cover the locations of interest, \eg, crack, hole, \etc~Based on this finding, we use the places of interest identified by the random CNN to synthesize pseudo abnormal patches in the pseudo anomaly generation module, which is the first attempt to use such inductive bias of random networks in 3D anomaly localization and detection.

In our proposed zero-shot 3D anomaly localization and detection setting, since the training data of the target objects are unavailable, we incorporate the 3D data from other objects (which we refer to as task-irrelevant data, \ie, the objects belonging to the categories different from the testing category) to synthesize the pseudo anomaly patches. We extract the 3D features of both the normal patches and pseudo abnormal patches and use a contrastive learning objective to further regularize the learned 3D feature extraction network. To enhance the anomaly localization and detection ability, we also introduce a normalcy classifier to distinguish the normal patches from the pseudo-abnormal patches to gain the discriminative ability between general normal and abnormal 3D objects. We add adversarial perturbations to the input point cloud patch utilizing the gradient of the negative log-likelihood loss applied to the testing data. Eventually, we combine the normalcy classification output score of the perturbed data and the distance-based score of the original using the RGB and FPFH features plus our learned 3D features to formulate the final anomaly score. We demonstrate that our proposed method \ours outclasses the SOTA 3D anomaly localization and detection method within the zero-shot framework. In summary, our key {\bf contributions} are as follows:
\begin{enumerate}
[topsep=0pt,itemsep=-1ex,partopsep=0ex,parsep=1ex,wide,labelwidth=!,labelindent=0pt]
    \item We formally introduce a new problem in 3D anomaly detection and localization where the model undergoes training using the normal data to detect anomalies (during testing) in a varied class without undergoing any adaptation through the target-class training data.
    \item We propose a novel zero-shot 3D anomaly detection and localization method, \ours, where our designed network learns the relative and general difference between the normal and abnormal 3D object data in the training class and generalizes to the target class without needing the target class training data or any models pre-trained by 3D data.
    \item Intriguingly and notably, for the very first time (as far as we are aware), we show that a randomly initialized and untrained CNN has the inductive bias to localize places of interest on three-dimensional point cloud data, and its localization ability is better than an ImageNet-pretrained CNN.
    \item As far as we are aware, this is the first attempt to incorporate the input perturbation technique into 3D anomaly detection and localization problems and show its efficacy.
\end{enumerate}

\section{Related works}
\noindent \textbf{3D anomaly localization and detection} is crucial in industrial scenarios~\cite{visapp22,Horwitz_2023_CVPR}. With the emergence of the first 3D anomaly localization and detection dataset MVTec 3D-AD~\cite{visapp22}, a great number of anomaly detection and localization methods for three-dimensional point cloud data have been introduced.~\cite{visapp22} proposed to use generative adversarial networks, autoencoders, and variational models in both voxel-level and depth-level modeling.~\cite{bergmann2023anomaly,rudolph2023asymmetric} adopted student-teacher frameworks for anomaly detection and take advantage of the distance between the student and teacher model output as anomaly score.~\cite{Horwitz_2023_CVPR} proposed the 3D version of Patchcore~\cite{roth2022towards}, which utilizes a core-set assisted memory bank for normal feature storage and employs the distance between the testing sample feature and the normal memory bank as an anomaly score. More recently,~\cite{cao2023collaborative} proposed a collaborative discrepancy optimization method with the help of synthetic anomalies,~\cite{wan2022position} came up with a new position-encoding-augmented feature mapping for anomaly detection, and~\cite{wang2023multimodal} suggested a hybrid feature fusion technique for multimodal industrial anomaly detection.~\cite{chu2023shape} developed a method using a dual-expert framework that combines 3D geometric information and 2D color features, but it required training the expert models using the training data of all categories, which is not feasible under our proposed zero-shot setting.~\cite{zavrtanik2024cheating} proposed a novel method called 3DSR, which utilizes a Depth-Aware Discrete Autoencoder architecture and a simulated depth data generation process to jointly model RGB and 3D data, achieving the best anomaly localization and detection performance so far. Despite the efficacy of the above solutions, they require the testing samples to share the same class as the training samples. Once the training and testing data distributions differ, the performance will be largely compromised. In contrast, \ours learns more generally discriminative features and aims to find the essential difference between normal and abnormal 3D data. Specifically, we propose to employ the inductive bias to generate pseudo-abnormal examples and use contrastive learning on top of them.

\begin{figure*}[tb]
  \centering
  \includegraphics[width=.98\linewidth]{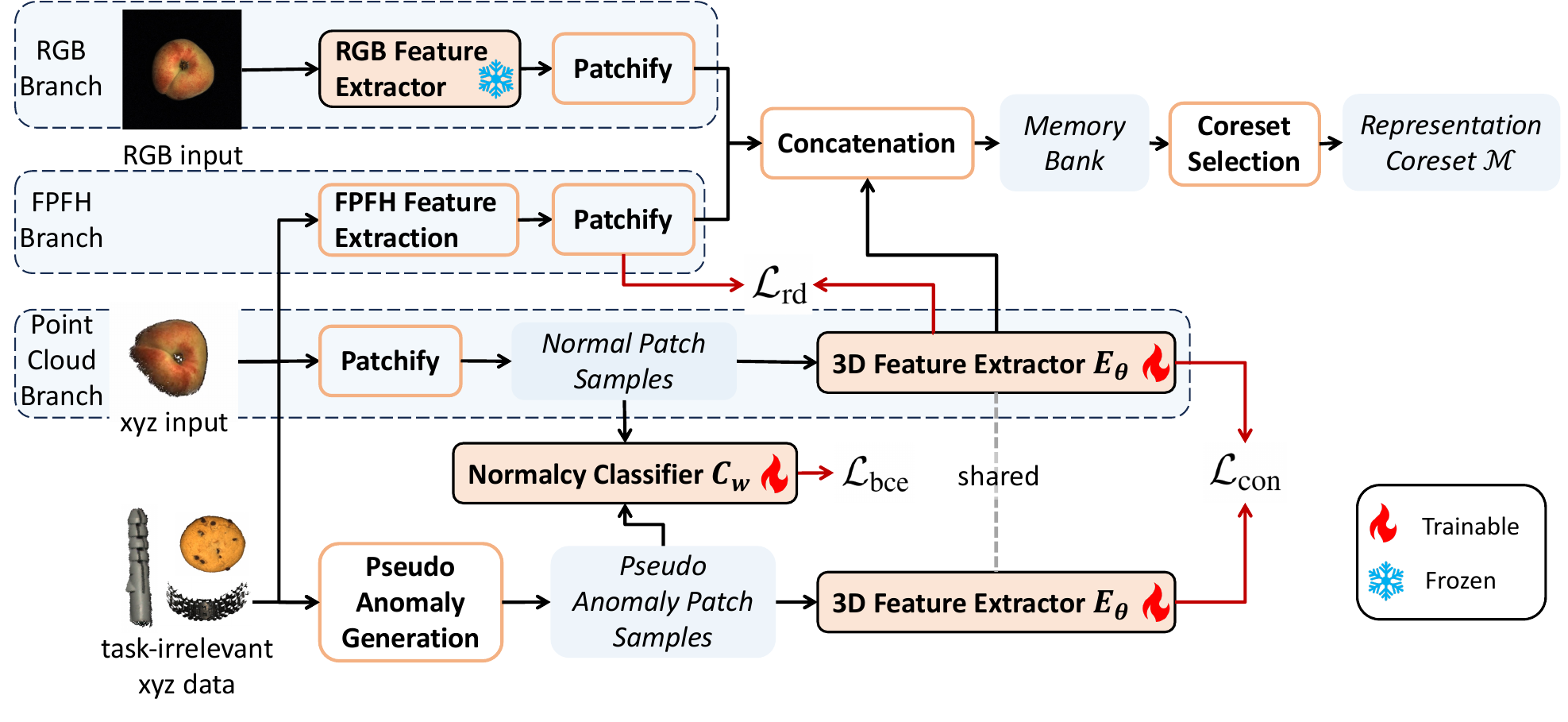}
  \caption{{\bf Framework overview.} Our proposed \ours framework mainly adopts three branches to extract features given both 2D and 3D data of an object. The RGB branch extracts feature from 2D image data of the object using ResNet pre-trained on ImageNet. The FPFH branch extracts handcrafted FPFH features from 3D point cloud data. The point cloud branch employs a learnable network (PointNet++) to extract features. The network is trained by a patch-level contrastive learning loss, which takes inductive bias-based pseudo anomaly patches as negative samples and normal patches as positive samples and a representation disentanglement loss which pushes the FPFH features and the learned 3D features away. The features of the three branches are concatenated to store in the memory bank where a coreset selection is performed. In addition, a normalcy classifier is trained to classify the pseudo anomaly patch and the normal patch using the binary cross-entropy loss.}
  \label{fig: framework}
  \vspace{-.7em}
\end{figure*}

\noindent \textbf{Low-shot anomaly detection} which is composed of few-shot anomaly detection and zero-shot anomaly detection, has been attracting attention in anomaly detection research recently. For few-shot anomaly detection, some works~\cite{sheynin2021hierarchical,huang2022registration,wu2021learning,defard2021padim,cohen2020sub,roth2022towards} reflected the notion of ``few-shot" in only using a much smaller number of normal training samples, and others~\cite{liznerski2021explainable,pang2021explainable,ding2022catching,lu2020few,ho2024long,venkataramanan2020attention} explored the setting that a few abnormal samples can be accessed during testing. 
In the context of zero-shot anomaly detection, the current dedication to such research direction is still limited.~\cite{esmaeilpour2022zero,liznerski2022exposing} exploited the transfer learning power of the pretrained CLIP models~\cite{radford2021learning} for image-level out-of-distribution detection or anomaly detection without the normal training data.~\cite{SCHWARTZ2024103958} investigated the capacity of ImageNet-pretrained masked autoencoder~\cite{he2022masked} for zero-shot image anomaly detection via adopting the reconstruction discrepancy as anomaly score, and~\cite{aich2023cross} tackled the zero-shot setting in video anomaly detection. More recently,~\cite{zhou2023anomalyclip} leverages text prompts that are not tied to specific objects, allowing it to identify general patterns of normality and abnormality, making it effective for zero-shot anomaly detection across different domains.~\cite{jeong2023winclip} relies on custom-designed text prompts to map image features to abnormal areas, utilizing CLIP’s capabilities for zero-shot anomaly recognition. Better than all the existing works, \ours needs no pre-trained model and makes the first attempt to execute zero-shot 3D anomaly detection and localization.   %

\section{Proposed \ours framework}

\noindent \textbf{Method overview.} Given the normal training data from one particular class, our aim is to learn the representation that can ideally transfer across different classes without the need for the normal data of the testing class. To achieve this goal, we introduce an innovative \ours framework which is depicted in Fig.~\ref{fig: framework}. \ours is built on the basis of a memory bank restoration and feature distance calculation paradigm. Specifically, \ours is composed of a random CNN-based pseudo 3D anomaly sample generation module with the assistance of task-irrelevant data, 3D feature contrastive learning using pseudo anomaly, and a 3D point cloud sample normalcy classifier trained using the normal training sample and the synthesized pseudo anomaly sample. Finally, the distance-based score using the contrastive-learned features and the  normalcy classifier output score using perturbed patch inputs are weighted and integrated to form the final anomaly score. 

Our paper focuses on the 3D anomaly location detection, so the directions of existing zero-shot AD works in RGB images are complementary to what we propose. We intentionally do not make use of any existing zero-shot AD work and show that our proposed method still outperforms the SOTA. If we use existing zero-shot AD works, then we won't be able to claim such novelty in our method. We leave the integration of zero-shot RGB anomaly detection techniques~\cite{esmaeilpour2022zero,liznerski2022exposing,SCHWARTZ2024103958,jeong2023winclip,zhou2023anomalyclip} for performance gain for future work.

\noindent \textbf{Notation.}
We denote a 3D point cloud sample as $X \in \RR^{N \times 3}$ and its ordered version as $\cX \in \RR^{H \times W \times 3}$, where $N$ is the number of points and $H, W$ is the height and width of the corresponding 2D ordered map, either 3D data or 2D RGB values and $H \times W = N$. Here ``ordered" means the 3D point cloud data are in the ordered 2D image form but the three-channel values of each pixel are xyz instead of RGB values. The object $\cX$ is partitioned into patches along the width and height. We denote each patch as $x$ and $\cX=\Box_x$, where $\Box$ refers to the practice of ``realigning" $x$ ``based on their respective spatial location" as defined in~\cite{roth2022towards}. The 3D representation extraction network is denoted as $E_{\theta}$ with parameter $\theta$, and the normalcy classifier network is represented as $C_w$ with parameter $w$.%

\subsection{Normal feature extraction} 
The Patchcore~\cite{roth2022towards} is one of the SOTA methods for 2D industrial anomaly localization and detection on the MVTec-2D dataset~\cite{bergmann2019mvtec}. BTF~\cite{Horwitz_2023_CVPR} is the 3D data version of the PatchCore and achieves the SOTA anomaly detection and localization result on the dataset MVTec-3D~\cite{visapp22}. Following these works, we also first extract features from the normal data samples and store them in the memory bank. In particular, we extract RGB features from the 2D RGB image of the 3D object and handcrafted 3D FPFH~\cite{rusu2009fast} features from the corresponding 3D point cloud sample. As illustrated in the point cloud branch of Fig.~\ref{fig: framework}, we add an additional 3D network to extract learnable features. The network is learned using contrastive learning and a feature disentanglement objective.

\subsection{Learning discriminative 3D representations}

\noindent \textbf{Pseudo anomaly generation with inductive bias}. To generate satisfactory anomaly detection and localization performance, we require our point cloud branch to reflect a clear distinction between the testing anomalies and the testing normal samples. However, considering that the training samples belong to different categories compared to the testing class within the zero-shot setting, if we want to regularize the training of $E_{\theta}$, we need to mimic the disparities between the normal and the abnormal 3D samples regardless of the class prior. This motivates us to synthesize pseudo anomalies and perform contrastive learning between the pseudo anomalies and the normal samples.~\cite{cao2022random} showed that a randomly initialized and untrained convolutional neural network (CNN) inherently possesses an inductive bias to focus on objects, \ie, even a randomly initialized CNN can generate biased activation maps towards objects of interest on a 2D image. Aich \etal~\cite{aich2023cross} are the first to utilize such inductive bias of an untrained CNN to extract objects from task-irrelevant data and attach to the normal data to synthesize 2D pseudo anomaly image samples.

In this work, we discover that {\bf such inductive bias also exists for 3D point cloud data}. More specifically, given an ordered point cloud data $\cX \in \RR^{H \times W \times 3}$, we employ an untrained ResNet-50 randomly initialized using the He initialization~\cite{he2015delving}, which is denoted as $R(\cdot)$. Here the channel values of the input are xyz location values instead of RGB values when feeding the inputs. We choose to use the reciprocal second, third, and fourth layer output of $ R(\cX) \in \RR^{d \times h \times w}$ to generate and fuse the resulting activation maps. Specifically, the output sizes of the reciprocal second, third, and fourth layers are $14 \times 14$, $28 \times 28$, and $56 \times 56$, respectively. We first sum the values of all the channels in the feature map and then normalize them to the range $[0,1]$. 
\begin{figure}[tb]
    \centering
    \includegraphics[width=\linewidth]{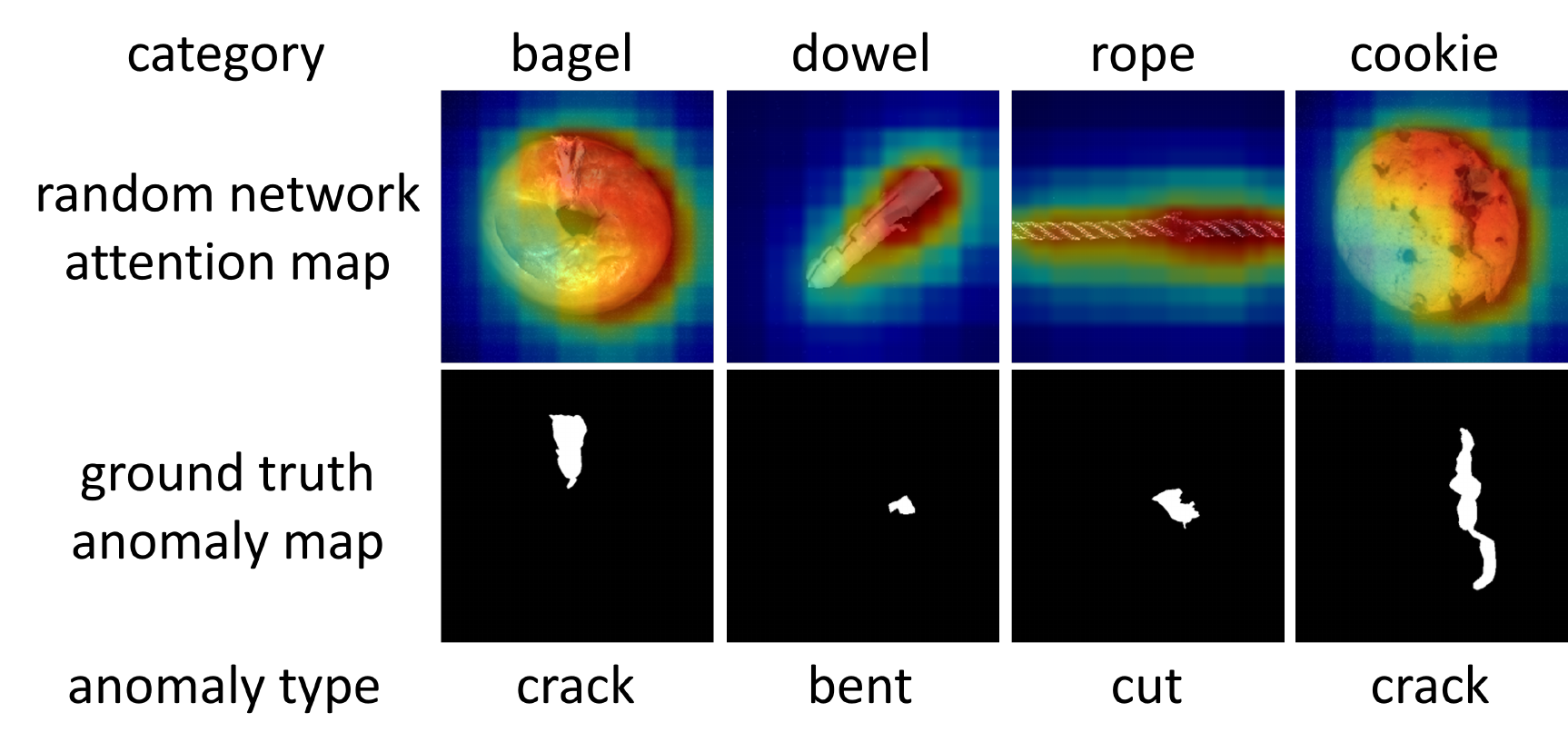}
    \caption{{\bf Inductive bias of random networks.} We feed the xyz data of abnormal examples as the input of a randomly initialized and untrained ResNet-50, and visualize the attention maps. These maps show that the random network has the inductive bias of covering the locations of interest, including the locations shown in the ground truth.}
    \label{fig: inductive bias}
\end{figure} 
Then we resize the output activation map of the reciprocal second and third layer output to the same spatial size as the reciprocal fourth layer, \ie, $56 \times 56$. Then the three activation map values are added, averaged, and resized to the original input size to generate the soft mask, which we dub as $A \in \RR^{H \times W}$. For the top $\tau$ (percentage) $A_{(i,j)}$ value points, we set mask $M_{(i,j)} = 1$, and for the rest, we set $M_{(i,j)}=0$.  Here $(i,j)$ represents the position information in $H\times W$ locations. Finally, the ordered points localized out are $M_X = M\odot \cX$, where $\odot$ is element-wise multiplication. As mentioned in~\cite{aich2023cross,cao2022random}, randomly initialized CNN is able to localize objects in that the background is comparitively less textured compared to the foreground object and the regions of foreground tend to exhibit higher activation values under activation functions like ReLU~\cite{hahnloser2000digital}. However, our method is different in that we use multi-scale attention values for information fusing, and also surprising because our fed input is ordered xyz tensor (the point location information), not RGB values.  In our experiments, we find that the highly activated areas correspond to the locations with shapes that are locations of interest (as illustrated in Fig.~\ref{fig: inductive bias}) and should be detected.  This is intriguing because it means that there also exists inductive bias for 3D xyz input data, which shows that at the spatial level, the point clusters that exhibit abrupt deviations or alternations can be highly activated under a series of activation functions like ReLU. After selecting the points that show the places of interest from the task-irrelevant data, we attach the points of interest to the normal training sample. Then we move our anchor point (the center point around which the points are selected or sampled) to the geometry center of the anomaly points part plus some surrounding points, and use KD-tree~\cite{zhou2008real} search algorithm to pick out the nearest point cloud part as the generated pseudo abnormal patch. The anomaly points part is attached to the surface by taking the anchor point as the geometric center. The anchor point serves as the query point in the KD tree search. The KD tree search is conducted within the normal point cloud plus the anomaly points part with the anchor point as the
query point. This can guarantee that the synthesized abnormal patch can contain both normal points and pseudo-generated points.

\begin{figure*}[tb]
    \centering
    \includegraphics[width=0.9\linewidth]{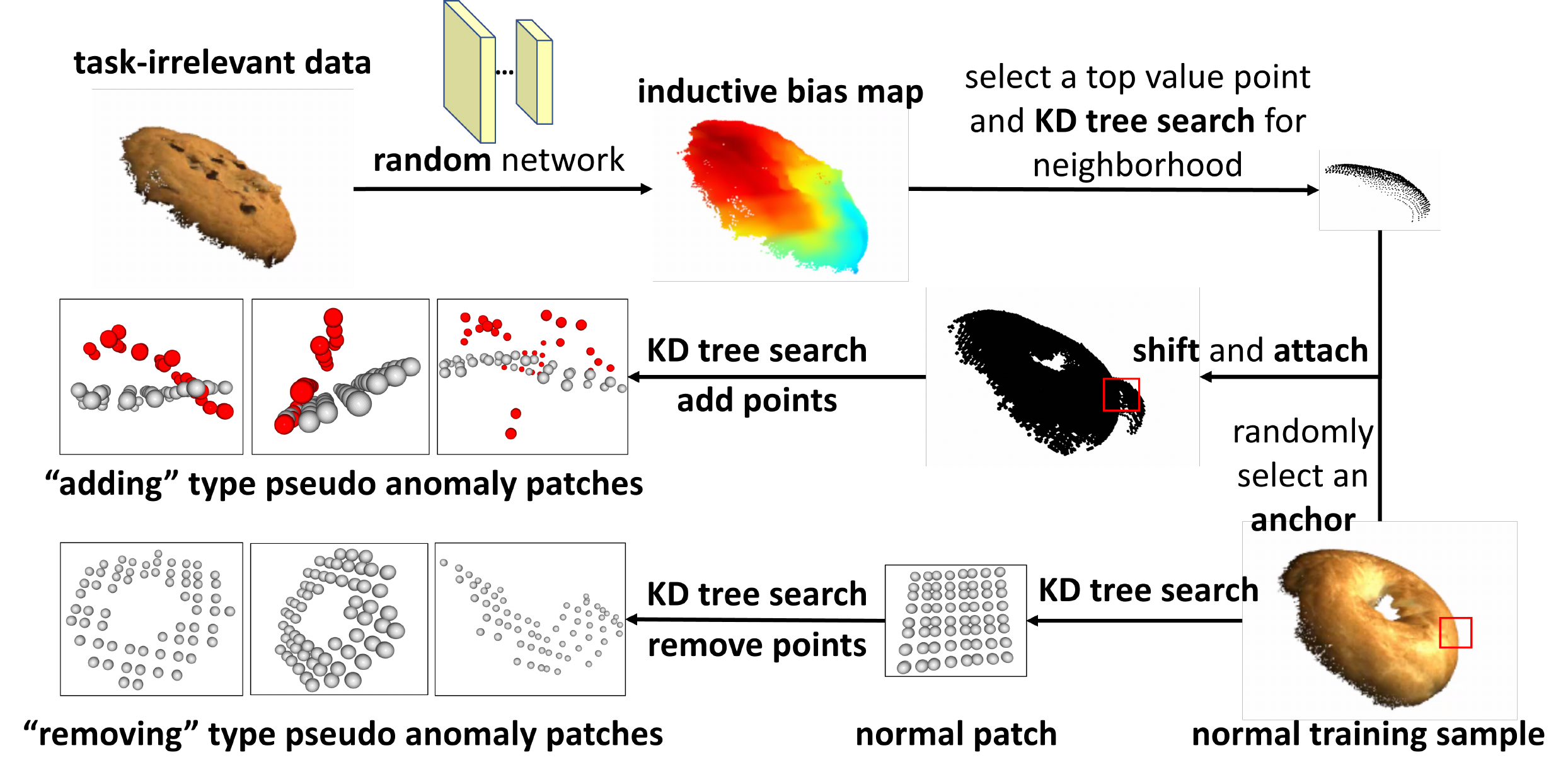}
    \caption{{\bf Pseudo anomaly generation.} Overview of our proposed patch-level 3D pseudo anomaly sample generation process for both ``adding" and ``removing" type anomalies.}
    \label{fig: pseudo-anomaly-generation}
\end{figure*}

Besides such ``adding-point" type pseudo abnormal patches which mimic anomaly types like bulging, contamination, or bent, we also involve another type of pseudo abnormal patch by setting the anchor at a random point of the surface of the normal sample, randomly sampling point cloud part and then randomly removing some ratio of points. Such kind of ``removing-point" anomaly aims to resemble abnormal part types including cuts or holes. Our generated pseudo abnormal patches consist of both ``adding-point" and ``removing-point" types with the quantity ratio $1:1$. Fig.~\ref{fig: pseudo-anomaly-generation} illustrates the above process. %

\noindent \textbf{Contrastive learning with pseudo anomalies}.
To learn the representations that can robustly distinguish between the intrinsically abnormal and normal 3D samples, we use a contrastive learning objective that takes normal 3D patches as positive samples and the pseudo-abnormal patches as negative samples. As shown in Fig.~\ref{fig: framework}, we add an additional 3D network PointNet++~\cite{qi2017pointnet++} $E_\theta$ in the patch level to extract features. We adopt the contrastive learning loss as: 
\fontsize{9.5pt}{10pt}
\begin{align}
    \cL_{\text{con}} = &\sum_{x_j \in \cX_p} \frac{-1}{|\cP(x_j)|} * \nonumber \\
    &\sum_{x_p \in \cP(x_j)} \log \frac{\exp\left(\frac{E_{\theta}(x_j) \cdot E_{\theta}(x_p)}{T\cdot \norm{E_\theta(x_j)}_2\cdot\norm{E_\theta(x_p)}_2}\right)}{\sum\limits_{x_n \in \cN(x_j)} \exp
    \left(\frac{E_{\theta}(x_j) \cdot E_{\theta}(x_n)}{T \cdot \norm{E_\theta(x_j)}_2\cdot \norm{E_\theta(x_n)}_2}\right)}, \label{eq: contrast}
\end{align}
\normalsize
where $\cX_p$ is the positive patch sample set, $\cP(x_j)$ is the positive patch sample set besides $x_j$, $\cN(x_j)$ is the negative patch sample set, and $T$ is the temperature parameter. The purpose of Eq.~\eqref{eq: contrast} is to maximize the similarity between the learned feature representations of the positive samples while minimizing the similarity between the positive sample set and the negative sample set. Since the positive sample patches exhibit normal patterns while the negative samples are pseudo abnormal patches which we use task-irrelevant data from multiple categories to generate, in the testing phase, the network learned by Eq.~\eqref{eq: contrast} has the capacity to induce features that are far away from the normal feature memory bank when encountering abnormal samples during testing. 

\noindent \textbf{Representation disentanglement loss}.
To ensure that the learned point cloud branch output features are complementary to the handcrafted FPFH features, we design the representation disentanglement loss $\cL_{\text{rd}}$, which aims at minimizing the cosine similarity $\text{cos}(\cdot)$ between the extracted learnable feature $E_{\theta}(x)$ and the FPFH feature $F(x)$:
\begin{equation}
    \cL_{\text{rd}} = \text{cos}(F(x),E_{\theta}(x)) = \frac{F(x)\cdot E_{\theta}(x)}{\norm{F(x)}_2\cdot\norm{E_{\theta}(x)}_2}.
\end{equation}
Therefore the loss function for training the network $E_{\theta}$ is the combination of the contrastive learning loss and the disentanglement loss: $\cL = w_{\text{con}} \cdot \cL_{\text{con}} + w_{\text{rd}} \cdot \cL_{\text{rd}}$, where $w_{\text{con}}$ and $w_{\text{rd}}$ are the weights.

\subsection{3D normalcy classifier}\label{sec: classifier}
For the positive and negative samples, we use an additional PointNet++~\cite{qi2017pointnet++} network $C_w$ for classification training. The normalcy classifier aims to distinguish between the normal sample and the synthetic abnormal ones and is formulated as a conventional binary classification problem. We adopt the binary cross-entropy loss $\cL_{\text{bce}}$:
\fontsize{9.5pt}{10pt}
\begin{equation}
    \cL_{\text{bce}} = - \frac{1}{N} \sum_{i=1}^N \log(p(x_i|w)) \cdot y_i + \log(1-p(x_i|w)) \cdot (1-y_i),
\end{equation}
\normalsize
where $x_i$s are the training data composed of positive (normal training sample patch) and negative (pseudo abnormal patch) samples of the contrastive learning paradigm, and $p(x_i|w)$ is the softmax output probability of class $1$ of $C_w$.  $y_i$s are binary labels and have value $0$ for positive samples and value $1$ for negative samples. In the testing phase, for each test sample patch, we use $p(x_i|w)$ as the patch-level anomaly score. The motivation is that we assign higher anomaly score values to the testing patch that is classified as abnormal (class $1$) and lower score values to the testing patch that is classified as normal (class $0$). This is because our classifier has been trained to discriminate between normal and pseudo-abnormal patches which are synthesized using task-irrelevant data belonging to multiple categories, which has been able to distinguish between the normal and abnormal 3D patches regardless of the class information.

\noindent \textbf{Training and memory bank}.
After the training of $E_{\theta}$, we extract the features of the training patches with it. The learned feature from the point cloud branch, concatenated with the RGB feature from RGB branch and the FPFH feature from FPFH branch, becomes the final feature representation of the patch $x$, and we denoted the concatenated feature of $x$ as $f(x)$. Inspired by PatchCore~\cite{roth2022towards}, we store the extracted features of the training samples into a memory bank and run a minimax facility location-based coreset selection~\cite{sener2018active,sinha2020small} algorithm to reduce the computation burden. We use notion $\cM$ for the reduced patch-level feature memory bank.

\subsection{Anomaly score design}\label{sec: score}
\noindent \textbf{Distance-based score}.
Given a testing object sample $X^{\text{test}}$, we extract the three branch patch-level features in the same way as the training process using trained $E_{\theta}$, and we denote the collected patch-level feature set as $f(\cX^{\text{test}})$ and the features as $f(x^{\text{test}})$. Following~\cite{roth2022towards}, we utilize the maximum distance score $S^*$ from $f(\cX^{\text{test}})$ to the corresponding nearest neighbour $f^*$ of the memory bank:
\begin{align}
f(x^{\text{test}, *}), f^* &= \argmax_{f(x^{\text{test}}) \in f(\cX^{\text{test}})} \argmin_{f \in \cM} \norm{f(x^{\text{test}}) - f}_2, \\
S^*&=\norm{f(x^{\text{test}, *}) - f^*}_2.
\end{align}
To obtain the object-level anomaly score $S_{\text{dist}}$, we impose an additional weight in the following form:
\begin{equation}
    S_{\text{dist}} (X^{\text{test}}) = \left(1-\frac{\exp\norm{f(x^{\text{test},*}) - f^*}_2}{\sum\limits_{f \in \cN_b(f^*)} \exp\norm{f(x^{\text{test},*}) - f}_2}\right)\cdot S^*,
\end{equation}
where $\cN_b(f^*)$ is the $b$ nearest patch-level features in the memory bank for the test patch-feature $f^*$. The rationale of adopting the reweighting strategy is that we ought to elevate the anomaly score when the nearest memory bank features to $f(x^{\text{test},*})$ and $f^*$ are themselves far away from the surrounding samples. To design the anomaly map score for pixel-level anomaly localization, we simply compute the $L_2$ distances between the test patch features and the nearest patch features in $\cM$, and realign them according to their respective spatial positions over the whole object. Then we resize the score map to the original ordered 3D data resolution $H \times W$ via Bilinear Interpolation $\cR$ and apply KNN Gaussian Blurring $\cB$ to the anomaly score map:
\fontsize{9.5pt}{10pt}
\begin{equation}
    S_{\text{dist,map}} (X^{\text{test}}) = \cB\left( \cR\left(\Box_{x^{\text{test}} \in \cX^{\text{test}} } \min_{f \in \cM} \norm{f(x^{\text{test}}) - f}_2\right)\right),
\end{equation}
\normalsize

\noindent \textbf{Classification-based score with 3D input perturbation}.
We apply adversarial perturbation to the input patch xyz point cloud data based on the gradient of the negative log of the softmax score from the anticipated class, as determined by our trained classifier $C_w$ in relation to the input patch. Mathematically, for any 3D point cloud patch $x^{\text{test}}$,
\begin{align}
    \tilde{x}^{\text{test}} = x^{\text{test}} + \eta (-\nabla_{x^{\text{test}}} \log (\hat{p}(x^{\text{test}}|w))),
\end{align}
where $\hat{p}(x^{\text{test}}|w) = \max\{p(x^{\text{test}}|w), 1-p(x^{\text{test}}|w)\}$, and $\eta$ is the perturbation magnitude. Given that $C_w$ has been effectively trained to classify between the normal and pseudo anomalous patch, this approach seeks to lower the softmax score of the class predicted with the highest likelihood. This means it aims to make the abnormality harder to categorize with respect to the testing sample. We denote the 3D object and its ordered version after perturbation as $\tilde{X}^{\text{test}}$ and $\tilde{\cX}^{\text{test}}$ respectively, \ie, $\tilde{\cX}^{\text{test}} = \Box_{x^{\text{test}} \in \cX^{\text{test}}}\tilde{x}^{\text{test}}$. We design the object-level classification-based score as 
\begin{equation}
    S_{\text{cls}}(\tilde{X}^{\text{test}}) =\max_{x^{\text{test}} \in \tilde{\cX}^{\text{test}}} p(\tilde{x}^{\text{test}}|w),
\end{equation}
meaning that the abnormality extent of the whole object is decided by the most abnormal patch considered by $C_w$. The classification-based anomaly score map is designed by realigning the patch-level softmax probabilities based on their overall spatial locations and applying the same interpolation and blurring techniques as in our distance-based score map:
\begin{equation}
    S_{\text{cls,map}}(\tilde{X}^{\text{test}}) =\cB\left(\cR\left(\Box_{x^{\text{test}} \in \tilde{\cX}^{\text{test}}} p(\tilde{x}^{\text{test}}|w)\right)\right).
\end{equation}

\noindent \textbf{Final anomaly score}.
To design the final anomaly score, we combine both the distance-based score and the classification-based score after the adversarial perturbabtion. We denote the weight of the distance-based score  as $w_d$ and the weight of the classification-based score as $w_c$. The final pixel-level anomaly score map is computed as:
\begin{equation}
    S_{\text{map}}(X^{\text{test}}) = w_d \cdot S_{\text{dist,map}}(X^{\text{test}})  
    + w_c\cdot S_{\text{cls,map}}(\tilde{X}^{\text{test}}),
\end{equation}
and the object-level final anomaly score is designed as 
\begin{equation}
     S(X^{\text{test}}) = w_d \cdot S_{\text{dist}}(X^{\text{test}}) + w_c\cdot S_{\text{cls}}(\tilde{X}^{\text{test}}).
\end{equation}

\vspace{-1em}
\section{Experiments}
\label{sec: exp}
\begin{table*}[tb]
\centering
\resizebox{\textwidth}{!}{
\begin{tabular}{cccccccccccccc}
\toprule
{train\textbackslash test} & bagel & cable & carrot & cookie & dowel & foam & peach & potato & rope & tire & mean (\ours) & BTF & 3DSR\\
\midrule
bagel & - & 78.6 & 91.9 & 89.3 & 81.6 & 48.3 & 91.2 & 96.5 & 82.0 & 86.7 & {\bf 82.9} (\ua{2.4}) &\underline{80.5} & 7.9 \\
cable & 31.5 & - & 87.1 & 48.6 & 81.0 & 56.6 & 65.4 & 89.4 & 81.8 & 80.7 & {\bf 69.1} (\ua{0.9}) &\underline{68.2} & 6.6 \\
carrot & 45.4 & 77.2 & - & 52.9& 82.3 & 46.3& 67.3& 91.3 & 84.4 & 89.3 & {\bf 70.7} (\ua{2.1}) &\underline{68.6} & 21.7\\
cookie & 70.6 & 76.8& 91.5 & - & 81.0 & 46.5 & 82.9 & 91.7 & 84.4 & 89.2 & {\bf 79.4} (\ua{5.5}) &\underline{73.9} & 8.3 \\
dowel & 15.1 & 76.7 & 89.8 & 20.8 & - & 46.4 & 49.0& 84.5 & 82.3 & 89.3 & {\bf 61.5} (\ua{4.1}) &\underline{57.4} & 35.4\\
foam & 25.6 & 77.6 & 86.0 & 9.4 & 80.1 & - & 57.0 & 79.4 & 79.8 & 83.9 & {\bf 64.3} (\ua{3.3}) &\underline{61.0} & 0.5 \\
peach & 81.3 & 78.8 & 92.4& 84.8 & 82.7& 51.8 &  -& 97.9 & 82.9& 89.2 & {\bf 82.4} (\ua{3.5}) &\underline{78.9} & 14.2\\
potato & 78.0& 78.1 & 96.7& 80.0& 81.5 & 46.4& 88.8& - & 82.7 & 88.1& {\bf 80.0} (\ua{1.7}) &\underline{78.3} & 13.2\\
rope & 13.4 & 76.3 & 87.7 & 9.0 & 80.5& 45.8& 47.7& 82.7 & - & 89.4 & {\bf 59.2} (\ua{7.2}) &\underline{52.0} & 19.3\\
tire & 14.9 & 76.7 & 87.8 & 6.5 & 80.8 & 47.0  & 48.4 & 83.7 & 80.8 & - & {\bf 58.5} (\ua{4.7}) & \underline{53.8} & 23.8\\
\bottomrule
\end{tabular}}
\caption{The detailed pixel-level AUPRO (\%) of \ours under the zero-shot setting. The best and second-best performances are highlighted in \textbf{bold} and \underline{underline} (the gain of \ours over the best baseline is also reported).~\ours outperforms BTF and 3DSR in all of the categories.}
\label{table: pixel-aupro}
\vspace{-.5em}

\end{table*}

\begin{table*}[tb]
\centering
\resizebox{\textwidth}{!}{
\begin{tabular}{cccccccccccccc}
\toprule
{train\textbackslash test} & bagel & cable & carrot & cookie & dowel & foam & peach & potato & rope & tire & mean (\ours) & BTF & 3DSR \\
\midrule
bagel & - & 57.9 & 71.8 & 68.9 & 57.5 & 58.1 & 56.1 & 69.0 & 47.3 & 55.9& {\bf 60.3} (\ua{6.9}) & \underline{53.4} & 46.1 \\
cable & 52.5 & - & 52.3 & 49.6 & 51.6 & 72.4 & 46.4 & 48.2 & 44.2 & 59.9 & {\bf 53.0} (\ua{2.9}) & \underline{50.1} & 47.7 \\
carrot & 51.6 & 54.1 & - & 53.5 & 57.9 & 54.9& 52.5& 48.8 & 48.1 & 47.1 & {\bf 52.1} (\ua{1.3}) & \underline{50.8} & 46.0 \\
cookie & 40.2 & 50.0 & 55.5 & - & 55.3 & 60.1 & 46.0 & 47.3 & 35.9 & 59.6 & \underline{50.0} (\da{0.6}) & 49.5 & {\bf 50.6} \\
dowel & 54.9 & 57.0 & 42.4 & 51.8 & - & 57.6 & 50.4& 55.9 & 58.9 & 50.4& {\bf 53.3} (\ua{2.0}) & \underline{51.3} & 47.3 \\
foam & 60.9 & 48.8 & 46.7 & 47.0 & 52.7 & - & 49.0& 48.5 & 50.0& 62.4& \underline{51.8} (\da{3.8}) & 49.9 & {\bf 55.6} \\
peach & 46.5 & 49.2 & 67.5 & 49.3& 55.0 & 54.8 & - & 79.7 & 58.1 & 52.6 & {\bf 57.0} (\ua{1.4}) &\underline{55.6} & 45.0 \\
potato & 43.8 & 50.5 & 73.8 & 43.2& 52.4& 55.6 & 49.7 & - & 46.2 & 48.3 & \underline{51.5} (\da{0.6}) &{\bf 52.1} & 49.7 \\
rope & 48.8 & 47.6 & 54.6 & 42.0 & 41.9 & 52.8 & 46.7 & 45.7&  -& 61.8& \underline{49.1} (\da{1.0}) &\underline{49.1} & {\bf 50.1} \\
tire & 48.0 & 52.1 & 48.8 & 45.5 & 51.6 & 63.5  & 53.6 & 50.0 & 56.3 & - & {\bf 52.2} (\ua{0.8}) & 50.5 & \underline{51.4} \\
\bottomrule
\end{tabular}}
\caption{The detailed image-level AUROC of \ours under the zero-shot setting. The best and second-best performances are highlighted in \textbf{bold} and \underline{underline} (the gain of \ours over the best baseline is also reported).~\ours outperforms BTF and 3DSR in most of the categories.}
\label{table: image-auroc}

\end{table*}

\noindent \textbf{Dataset}.
We conduct our zero-shot setting experiments on the MVTec 3D-AD dataset~\cite{visapp22}, which is the most commonly used 3D anomaly detection and localization dataset for industrial inspection. The dataset MVTec 3D-AD is a collection of high-resolution 3D models and corresponding 2D images. The dataset contains more than 800 3D models of everyday objects from 10 different classes.

\noindent \textbf{Experimental setting}.
For our proposed zero-shot setting, we iteratively use the normal training data of one class for the training of \ours, and then test on a different class. To ensure that the auxiliary data for pseudo anomaly generation is task-irrelevant, we adopt the leave-one-out strategy. Specifically, we use one class chosen from the remaining $9$ classes for testing and the rest $8$ classes as the task-irrelevant data in turn. There are 10$\times$9=90 individual experiments in total. 

\noindent \textbf{Implementation details}.
For the RGB branch feature extraction, we use the Wide ResNet-50~\cite{Zagoruyko2016WRN} pre-trained on the ImageNet~\cite{deng2009imagenet}. For all the ordered 3D data, we resize the original data resolution to $H$=$W$=224 and use the 8$\times$8 patch size, so for each sample, we have 28$\times$28=784 patches. For both $E_{\theta}$ and $C_w$ we adapt the input resolution of PointNet++~\cite{qi2017pointnet++} network architecture to $64$. We use the percentage $\tau$=0.1\% when choosing pseudo abnormal points, and we choose the ratio of negative patch samples and positive patch samples as $16:1$ for contrastive learning. We set the temperature $T$=0.07 in our contrastive learning paradigm. We set the weights of the loss functions as $w_{\text{con}}$=1, $w_{\text{rd}}$=100 to make the range of each loss comparable. The Adam optimizer~\cite{kingma2014adam} is employed for training. We train the two networks $E_{\theta}$ and $C_w$ for $5$ epochs and use the last-epoch model in the testing phase. At testing time, we set the nearest neighbor parameter $b$=3 and the perturbation magnitude $\eta$=0.1. 

\noindent \textbf{Baselines and evaluation metric}.
Since the problem of ``zero-shot 3D anomaly localization" is defined by us, we are unable to identify alternative methods specifically designed for this setup. The most recent and closely related baselines we find are BTF~\cite{Horwitz_2023_CVPR} and 3DSR~\cite{zavrtanik2024cheating}, which work under the classical setting, \ie, the training and the testing class are the same. We adopt the commonly used evaluation metrics: pixel-level AUPRO and image-level AUROC~\cite{visapp22,Horwitz_2023_CVPR,zavrtanik2024cheating}. 3DSR is the current best SOTA work under the classical setting, which achieves the highest image-level AUROC up to $0.978$ and the highest pixel-level AUPRO up to $0.972$ (mean taken over all the classes in the classical setting). We adapt the publicly released code of 3DSR~\cite{3dsr_github} and BTF~\cite{btf_github} to our zero-shot setting to report their results respectively for a fair comparison.

\noindent \textbf{Result and analysis}.
We summarize the pixel-level AUPRO of \ours and compare them with the mean results of BTF and 3DSR in Tab.~\ref{table: pixel-aupro}. Tab.~\ref{table: pixel-aupro} shows that \ours outperforms BTF/3DSR in all the categories by a considerable margin, which shows the efficacy of \ours. It also outperforms BTF/3DSR in {\bf 9/7 out of 10} categories in image-level AUROC, as shown in Tab.~\ref{table: image-auroc}. Despite performing well in the classic setting, 3DSR performs particularly badly in anomaly localization in our zero-shot setting, which shows its poor generalization ability. %

\noindent \textbf{Comparison of the memory bank size and the model parameters size.} As to the memory bank size after coreset selection, when trained on the class bagel and testing on the class cable gland, the size of the memory bank of baseline BTF is 229M (19129 patch features with dimension size 1569) and that of \ours is 234M (19129 patch features with dimension size 1601). Therefore, \ours has a slightly larger memory bank than BTF (because we have additionally concatenated learned 3D features for each memory bank patch-level feature), and the size ratio is the same for other training and testing cases. Although the memory bank size of \ours increases 2.2\% compared with BTF, the pixel-level AUPRO of \ours improves over BTF by about 5.7\% on average. For the model parameters, the model parameter size of 3DSR is 38M, and the model size of \ours is 13.6M (6.8M for the learned 3D feature network and 6.8M for the normalcy classifier). Therefore, \ours has nearly $\frac13$ of the model parameters of 3DSR. BTF method does not involve model training.

\begin{table}[t]
\centering
\resizebox{\columnwidth}{!}{
\begin{tabular}{@{}c@{\hspace{2mm}}c@{\hspace{2mm}}c@{\hspace{2mm}}c@{\hspace{2mm}}c@{\hspace{2mm}}c@{\hspace{3mm}}c@{}}
   \toprule
    baseline &$\cL_{\text{con}}$ &$\cL_{\text{rd}}$  &$C_w$   &IP &P-AUPRO (\%) &I-AUROC (\%) \\
    \midrule
    \ccheck &\ccross &\ccross &\ccross &\ccross &80.5 / 57.4 / 61.0 &53.4 / 51.3 / 49.9 \\
    \ccheck &\ccheck &\ccross &\ccross &\ccross & 80.6 / 57.7 / 62.0 &53.7 / 51.5 / 50.2\\
    \ccheck &\ccheck &\ccheck &\ccross &\ccross &  80.8 / 57.8 / 62.5 &54.0 / 51.8 / 50.5\\
    \ccheck &\ccheck &\ccheck &\ccheck &\ccross & 82.7 / 61.3 / 63.8 &57.5 / 52.4 / 50.9\\
    \ccheck &\ccheck &\ccheck &\ccheck &\ccheck &  \textbf{82.9} / \textbf{61.5} / \textbf{64.3} &\textbf{60.3} / \textbf{53.3} / \textbf{51.8}
\\
   \bottomrule
   \end{tabular}
}
\caption{Ablation study of the \ours components. IP denotes input perturbation, P-AUPRO denotes pixel-level AUPRO, and I-AUROC denotes image-level AUROC. For each cell, the numbers correspond to the cases when the training
class is bagel/dowel/foam.}
\vspace{-1em}
\label{table: ablation-component}
\end{table}

\begin{table}[t]
\centering
\resizebox{\columnwidth}{!}{
   \begin{tabular}{@{}c@{}c@{}cc@{}}
   \toprule
    \multicolumn{2}{c}{pseudo anomaly generation type} &pixel-level &image-level  \\
    \cmidrule(lr){1-2}
    adding points & removing points &AUPRO (\%) &AUROC (\%)  \\
    \midrule
    \ccheck &\ccross & 80.3 / 79.3 / 58.8  & 53.8 / 52.6 / 49.1\\
    \ccross &\ccheck & 81.0 / 79.6 / 58.9 & 54.0 / 52.8 / 49.1\\
    \ccheck &\ccheck &  \textbf{82.9} / \textbf{80.0} / \textbf{59.2} & \textbf{60.3} / \textbf{51.5} / \textbf{49.1} \\
    \bottomrule
   \end{tabular}
}
\caption{Ablation study of the pseudo anomaly generation type. For each cell, the numbers correspond to the cases when the training class is bagel/potato/rope.}
\vspace{-1em}
\label{table: ablation-pseudo}
\end{table}

\noindent \textbf{Ablation study}.
We carry out ablation studies on the components of \ours including the loss functions, normalcy classifier, and input perturbation. We conduct experiments on the 3 diverse training categories: bagel (round and relatively big), dowel (strip-shaped), and foam (irregular shape). Tab.~\ref{table: ablation-component} shows that $\cL_{\text{con}}, \cL_{\text{rd}}$, $C_w$, and input perturbation all enhance the performance. 
Next, we study the impact of the patch-level pseudo anomaly type on the performance by getting rid of the ``adding-point" and ``removing-point" type anomalies, and summarize the results in Tab.~\ref{table: ablation-pseudo}, where both pseudo anomaly types contribute to 3D anomaly localization performance. 
Moreover, we use the ResNet-50 model pre-trained on the ImageNet dataset instead of random initialization in Tab.~\ref{table: ablation-pretrained}, where we conduct experiments to train on the class bagel/peach/rope, and all the other experimental settings and hyperparameters are kept the same. Tab.~\ref{table: ablation-pretrained} shows that random initialization outperforms the pre-trained one for pseudo anomaly generation in our task. We hypothesize that it is because the ImageNet pretrained weights focus on discriminative areas for classification purposes, not necessarily the abnormal areas we want. %
Finally, we show the experiments when training on 2 classes and testing on another unseen class for both BTF and \ours~in Tab.~\ref{table: multi-normal}, where \ours~consistently outperforms BTF in anomaly localization, which supports that \ours~can generalize to the multi-class setting.

\begin{table}[t]
\centering
\resizebox{\columnwidth}{!}{
   \begin{tabular}{@{}c@{\hspace{2.5mm}}c@{\hspace{2.5mm}}c@{}}
   \toprule
   \multirow{2}{*}{method} &pixel-level &image-level\\
    &AUPRO (\%) &AUROC (\%)\\
   \midrule
   BTF (baseline) &80.5 / 78.9 / 52.0 &53.4 / 55.6 / 49.1\\
   \ours~(pretrained CNN WI) & 81.4 / 79.6 / 57.8 &56.7 / 56.0 / 49.1\\
   \ours~(random CNN WI) &\textbf{82.9} / \textbf{82.4} / \textbf{59.2} &\textbf{60.3} / \textbf{57.0} / \textbf{49.1}\\
   \bottomrule
   \end{tabular}
}
\caption{Ablation study on the CNN weight initialization (WI) type. For each cell, the numbers correspond to the cases when the training class is bagel/peach/rope.}
\vspace{-1em}
\label{table: ablation-pretrained}
\end{table}

\begin{table}[t]
\centering
\resizebox{\columnwidth}{!}{
   \begin{tabular}{@{}c@{\hspace{1mm}}cc@{}}
   \toprule
   training classes $\backslash$ method &BTF &\ours\\
   \midrule
   bagel + cable & 80.3 / 53.4 &{\bf 80.9} (\ua{0.6}) / {\bf 53.7} (\ua{0.3}) \\
   carrot + cookie & 79.4 / 50.8 &{\bf 83.7} (\ua{4.3}) / {\bf 55.4} (\ua{4.6}) \\
   dowel + foam & 78.5 / 51.3 &{\bf 80.3} (\ua{1.8}) / {\bf 54.0} (\ua{2.7})\\
   \bottomrule
   \end{tabular}
}
\caption{The performance of training on the normal data of the specified 2 classes, and testing on the rope class. For each cell, the first / second number is pixel-level AUPRO (\%) / image-level AUROC(\%). We also report the performance gain of \ours over BTF.}
\vspace{-1em}
\label{table: multi-normal}
\end{table}

\vspace{-.5em}
\section{Conclusion and Limitation}
\label{sec: conclusion-impact-limitation}
We have defined a new task for 3D anomaly localization and detection, which involves localizing anomalies in 3D point clouds for the target class that lacks training data. To address this challenge, we have proposed a novel framework named ``3D zero-shot anomaly localization" (\ours) that aims to learn patch-level relative normalcy using contrastive learning and normalcy classification based on pseudo abnormal 3D patch generation. We are the first to show the efficacy of input perturbation in 3D anomaly detection and localization. \ours~surpasses the current state-of-the-art methods in 3D anomaly detection and localization. These promising results highlight the potential of using task-irrelevant data to generate pseudo anomalies as a viable approach for tackling the zero-shot 3D anomaly detection and localization problem. In addition, our new finding that a randomly initialized untrained neural network has the inductive bias to localize places of interest on 3D data can be potentially utilized as a prior for other tasks involving 3D data. The anomaly localization performance in some cases (\eg, train on the class foam and test on the class cookie) is not high.

\clearpage
{\small
\bibliographystyle{ieee_fullname}
\bibliography{zs3dad}
}

\end{document}